\documentclass[preprint,12pt]{elsarticle}
\usepackage[utf8]{inputenc}
\usepackage[T1]{fontenc}
\usepackage{graphicx}
\usepackage{multirow}
\usepackage{subcaption}
\usepackage{pdfpages}
\usepackage{lscape}
\usepackage{xcolor}
\usepackage{sectsty}

\subsectionfont{\normalfont\bfseries}
\usepackage{fancyhdr}
\usepackage{fancybox}
\usepackage{longtable}
\usepackage{amsmath,amssymb,amsfonts}
\usepackage{graphicx}
\usepackage{multirow}
\usepackage{caption}
\usepackage{subcaption}
\usepackage{algorithmicx}
\usepackage{float}
\usepackage{makecell}
\usepackage{subcaption}
\usepackage{makecell}
\usepackage{tabularray}
\usepackage{array}
\usepackage{amsmath}
\usepackage{algorithm}
\usepackage[noend]{algpseudocode}
\usepackage[bottom]{footmisc}
\usepackage{amsmath,eqparbox}

\usepackage{amssymb}

\journal{Computer Methods and Programs in Biomedicine}

\begin{document}

\begin{frontmatter}

\title{Siamese Content-based Search Engine for a More Transparent Skin and Breast Cancer Diagnosis through Histological Imaging }

\author[inst1,inst2]{Zahra Tabatabaei}
\author[inst2]{Adrián Colomer}
\author[inst1]{Javier Oliver Moll}
\author[inst2]{Valery Naranjo}

\affiliation[inst1]{organization={Dept. of Artificial Intelligence},
            addressline={Tyris Tech S.L.}, 
            city={Valencia},
            postcode={46021}, 
            state={Valencia},
            country={Spain}}

\affiliation[inst2]{organization={Instituto Universitario de Investigación en Tecnología Centrada en el Ser Humano, HUMAN-tech },
            addressline={Universitat Politècnica de València}, 
            city={Valencia},
            postcode={46022}, 
            state={Valencia},
            country={Spain}}

\begin{abstract}
Computer Aid Diagnosis (CAD) has developed digital pathology with Deep Learning (DL)-based tools to assist pathologists in decision-making. Content-Based Histopathological Image Retrieval (CBHIR) is a novel tool to seek highly correlated patches in terms of similarity in histopathological features. In this work, we proposed two CBHIR approaches on breast (Breast-twins) and skin cancer (Skin-twins) data sets for robust and accurate patch-level retrieval, integrating a custom-built Siamese network as a feature extractor. The proposed Siamese network is able to generalize for unseen images by focusing on the similar histopathological features of the input pairs. The proposed CBHIR approaches are evaluated on the Breast (public) and Skin (private) data sets with top \textit{K} accuracy. Finding the optimum amount of \textit{K} is challenging, but also, as much as \textit{K} increases, the dissimilarity between the query and the returned images increases which might mislead the pathologists. To the best of the author's belief, this paper is tackling this issue for the first time on histopathological images by evaluating the top first retrieved images. The Breast-twins model achieves 70\% of the F1score at the top first, which exceeds the other state-of-the-art methods at a higher amount of \textit{K} such as 5 and 400. Skin-twins overpasses the recently proposed Convolutional Auto Encoder (CAE) by 67\%, increasing the precision. Besides, the Skin-twins model tackles the challenges of Spitzoid Tumors of Uncertain Malignant Potential (STUMP) to assist pathologists with retrieving top \textit{K} images and their corresponding labels. So, this approach can offer a more explainable CAD tool to pathologists in terms of transparency, trustworthiness, or reliability among other characteristics.
\end{abstract}


\end{frontmatter}

\section{Introduction}
Skin cancer is one out of three diagnosed cancers worldwide, according to the World Health Organization (WHO) \cite{apalla2017epidemiological}. Basal and squamous cell carcinoma are the two most frequently occurring types of skin cancer, while the most perilous type is malignant \cite{reichrath2014epidemiology}. Classifying and distinguishing between benign (melanocytic nevus) and malignant (melanoma) can be reliably feasible in common melanocytic tumors \cite{soengas2003apoptosis}. One of the diagnostic challenges for pathologists in spitzoid tumors is Spitzoid Tumors of Uncertain Malignant Potential (STUMP) because their prognostic implications are unknown \cite{wiesner2016genomic}. In this case, to make an accurate cancer diagnosis, pathologists often need to transfer the biopsy glass to other centers to consult with their peers about the tissue grade. This workflow is time-consuming and expensive, and accidents might occur with the biopsy glass \cite{tabatabaei2023wwfedcbmir}.

Another world's most prevalent cancer type according to WHO is breast cancer with $685,000$ deaths globally in 2020 \cite{apalla2017epidemiological}. Breast cancer is caused by abnormal breast cells that grow out of control and form tumors. Cancer cells can spread to nearby lymph nodes or other organs. Treatment for breast cancer depends on the type and sub-type of cancer and how it has spread outside of the breast \cite{agbley2023federated}. There are some treatments for breast cancer including, surgery to remove the breast tumor, radiation therapy to reduce recurrence risk in the breast and surrounding tissues, and medications to kill cancer cells and prevent spread, including hormonal therapies, chemotherapy, or targeted biological therapies. Doctors might combine some treatments to make sure that the possibility of the cancer coming back is minimal. But the over-diagnosis or over-treatment, and variability in interpretation are some key challenges in breast cancer diagnosis \cite{huang2011time}.

Computer Aid Diagnosis (CAD) offers some efficient Deep Learning (DL) tools that can assist pathologists and address human errors in cancer diagnosis and treatment \cite{ahmedt2022survey}. The advance of DL methods opened the door to reducing the workload for pathologists and enhanced patient care \cite{Neel}. Digitization of the tissue slides as high-resolution Whole Slide Images (WSIs), emerges computer vision tools as a support for pathologists in their daily tasks. Some of these tools, such as classification \cite{shahdoosti2020object}, segmentation \cite{shahdoosti2019mri}, Content-Based Histopathological Image Retrieval (CBHIR) \cite{owais2019effective}, etc., provide a second opinion for pathologists to have a more accurate cancer diagnosis \cite{zahra_eusipco}. Image-based tools of CAD are classified into medical image classification and Content-Based Medical Image Retrieval (CBMIR) \cite{LEE20155356}. The main objective of the classifiers is to categorize the images, while CBMIR aims to rank and show similar images in addition to their labels \cite{LI2021675}. CBMIR, particularly CBHIR tools, demand representative features to retrieve similar histological patches with the same histopathological patterns effectively. To reach the most meaningful features of the patches, a proper Feature Extractor (FE) is needed.

Finding an optimum FE for the goal under study is challenging. There are plenty of FE, including SIFT \cite{zhang2014study}, color histogram-based features \cite{ma2018generating}, Gabor features \cite{ma2016breast}, and GIST features \cite{shi2017supervised} as some hand-craft features. In the other works \cite{mahraban2018transferred, jiang2016scalable, garg2021novel, singh2023joint}, DL-based FEs, including CNN, pre-trained models, Auto Encoder (AE), etc., were applied to extract the deep features. However, most of DL-based techniques face some challenges, mostly related to the digitized histopathological images. For instance, the annotated data set is one of the critical requirements for developing a high-performance DL- model. However, annotating histopathological images is time-consuming and costly, hindering the performance of DL-based tools. To cope with this challenge, some of the recent research on WSIs has focused on self/unsupervised techniques, which need fewer annotated images \cite{campanella2019clinical}.

As a means to tackle this lack of enough annotated images, Zahra et al. proposed an unsupervised ResCAE to perform a search engine in prostate cancer. This paper reported $85\%$, $78\%$ of accuracy at top 7 and 5, respectively \cite{Zahra_IVMSP}. Histopathology Siamese Deep Hashing (HSDH) is proposed in \cite{alizade} for histopathology retrieval and achieved $97\%, 98\%,$ and $99\%$ Mean Average Precision (MAP) for 32, 64, and 128 bits. The authors reported the results at the top 100, 150,..., and 400 for BreaKHis as a binary data set. RetCCL in \cite{wang2023retccl} is a clustering-guided contrastive learning approach for WSI retrieval. RetCCL has improved $24\%$ at the patch-level retrieval on the TissueNet data set in terms of $mMV@5$ in comparison with ImageNet pre-trained features. Authors in \cite{zheng2017size} proposed a CBHIR framework for a data set containing WSIs and size-scalable query Regions Of Interest (ROI). The reported retrieval results at the top 20 returned images, reached $96\%$ of precision on the Motic database. SMILY \cite{SMILY} was proposed by Google AI Healthcare and it used an automatic high-level feature extraction to provide the feature vectors for the search engine. SMILY retrieved images with the top 5 similar patches of prostate cancer with 73\% accuracy.

In the recent papers on CBHIR \cite{ tabatabaei2023wwfedcbmir, Zahra_IVMSP, alizade, zheng2017size, SMILY, tabatabaei2023towards, kalra2020yottixel}, the performance of the model was reported by top  \textit{K} accuracy. This evaluation technique assumes that if just one of the retrieved images out of \textit{K} is correct, the model performed correctly \cite{wang2023retccl}. 
Choosing the correct amount of \textit{K} in this evaluation technique is challenging. In some papers such as \cite{alizade}, the amount of \textit{K} was chosen in $[100-400]$, which is a high amount, especially for a binary data set. In \cite{zheng2017size}, the reported accuracy is at the top 20 images. In other papers \cite{hemati2023learning,kayhan2021content,majhi2021image,yu2021weber}, mostly the amount of \textit{K} was chosen as $5, 7, 10$. Retrieving top \textit{K} images brings some benefits for pathologists; for instance, they can analyze more cases to see similar histopathological features. The issue with top \textit{K} accuracy in evaluating the performance of CBHIR is that, in some cases, the model could retrieve one correct similar patch out of 5, 7, or even 400 retrieved images. Although this can ensure pathologists that at least one image out of \textit{K} is correct, it cannot provide a highly accurate second opinion for them. However, in the top \textit{K} accuracy technique, even if the $k_{th}$ retrieved out of \textit{K}, was the only similar image to the query, the accuracy of the model is high. This issue is highlighted more in the papers with a high amount of \textit{K}.

In a departure from conventional methods, this paper proposes a CBHIR by applying a custom-built Siamese network as an FE on skin and breast cancer data sets, which are the prevalent cancer types. Since these types of data sets can exhibit substantial variability, the Siamese network is a promising option for histopathological images. This network is adapted to capture the intricate histopathological patterns in the patches that are crucial for grading and diagnosis. The Siamese network is specifically designed to excel in similarity-based tasks and this feature is particularly well-suited for CBHIR applications. The proposed Siamese network employs a contrastive loss function to emphasize the relationships between data points in the embedding space. This behavior can enhance the ability of the proposed Siamese network to identify similarities between images and improve the distinction between dissimilar ones. The extracted features by the proposed Siamese network is highly transferable and they can be employed effectively in the search engine of CBHIR without extensive retraining. This transferability is valuable when dealing with histopathological images.

To the best of the author's knowledge, no previous studies were conducted based on the Siamese network in a CBHIR model to assist pathologists in making a more accurate diagnosis of spitzoid cancer. Also, this is the first time that a CBHIR model has been dedicated to grading the STUMP cases in a spitzoid melanocytic lesions database. Furthermore, in this paper, the proposed CBHIR model can have high performance with the top first retrieved patches, which is the first paper that could reach this power at the top first, to the author's best belief. The performance of the proposed CBHIR model confirms that the model is generalized well to unseen data.

In summary, this paper makes the following main contributions:
\begin{enumerate}
    \item We propose two Siamese networks for breast cancer (public data set) and skin cancer (private data set) to show the generalization of the proposed CBHIR technique. These proposed Siamese networks are robust to imbalanced data sets;
    \item Siamese network is employed to address the shortcomings of histopathological images, including small inter-class variations and large intra-class variances;
    \item The CBHIR results on both data sets were reported at the top first retrieved images to demonstrate the model's efficacy in retrieving relevant patches;
    \item The proposed CBHIR approach provides a second opinion to pathologists to tackle the challenges in grading STUMP by providing deep insights into the complexities. We show the Gradient-weighted Class Activation Mapping (Grad-CAM) figures as explainable Skin-twins to provide interpretability to the uncifrable STUMP cases;

    \item The performance of the proposed CBHIR technique in retrieving the images with the same cancer type was evaluated in comparison with some state-of-the-art classifiers;
    \item Based on the experimental results on two histopathological data sets, the proposed CBHIR framework, thanks to the Siamese network, outperforms other image retrieval methods.
\end{enumerate}

\section{Methodology}
\begin{figure*}[htp!]
\begin{center}
\centerline{\includegraphics[width=0.99\textwidth]{ 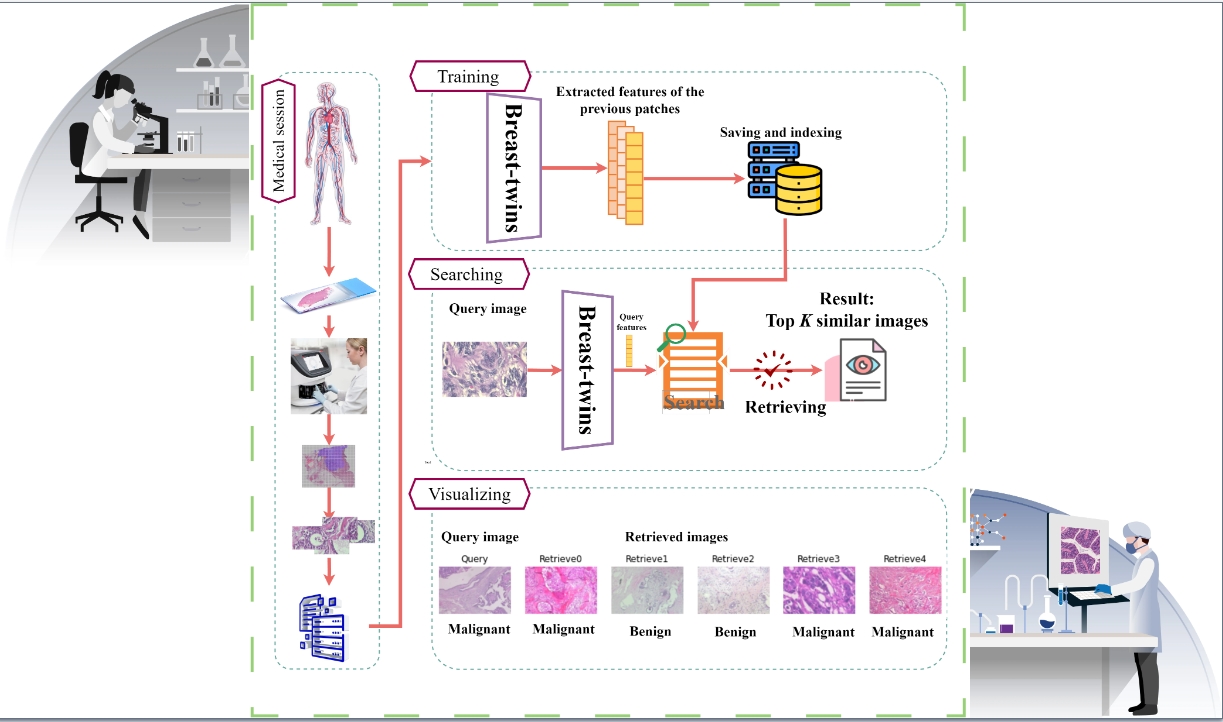}}
\end{center} 
\vspace{-0.7cm}
\caption{ A general overview of the proposed CBHIR to show how the proposed CBHIR can bridge traditional cancer diagnosis workflow to digital pathology.}
\label{fig:flowchart}
\end{figure*}

Figure \ref{fig:flowchart} illustrates the proposed CBHIR system in a general overview. First, medical centers have to scan the biopsies with WSI scanners \cite{racoceanu2015towards}. Second, the Siamese network needs to train to extract the features of the images. Third, the extracted features of the database have to be indexed and saved. Later, a search process needs to find similar patches to the query among the images in the database. Finally, it is time to visualize the top similar patches to pathologists for further analysis. This is how the CBHIR system can make a bridge between the current cancer diagnosis and digital pathology.
\subsection{Siamese network}
\begin{figure*}[htp!]
\begin{center}
\centerline{\includegraphics[width=0.73\textwidth]{ 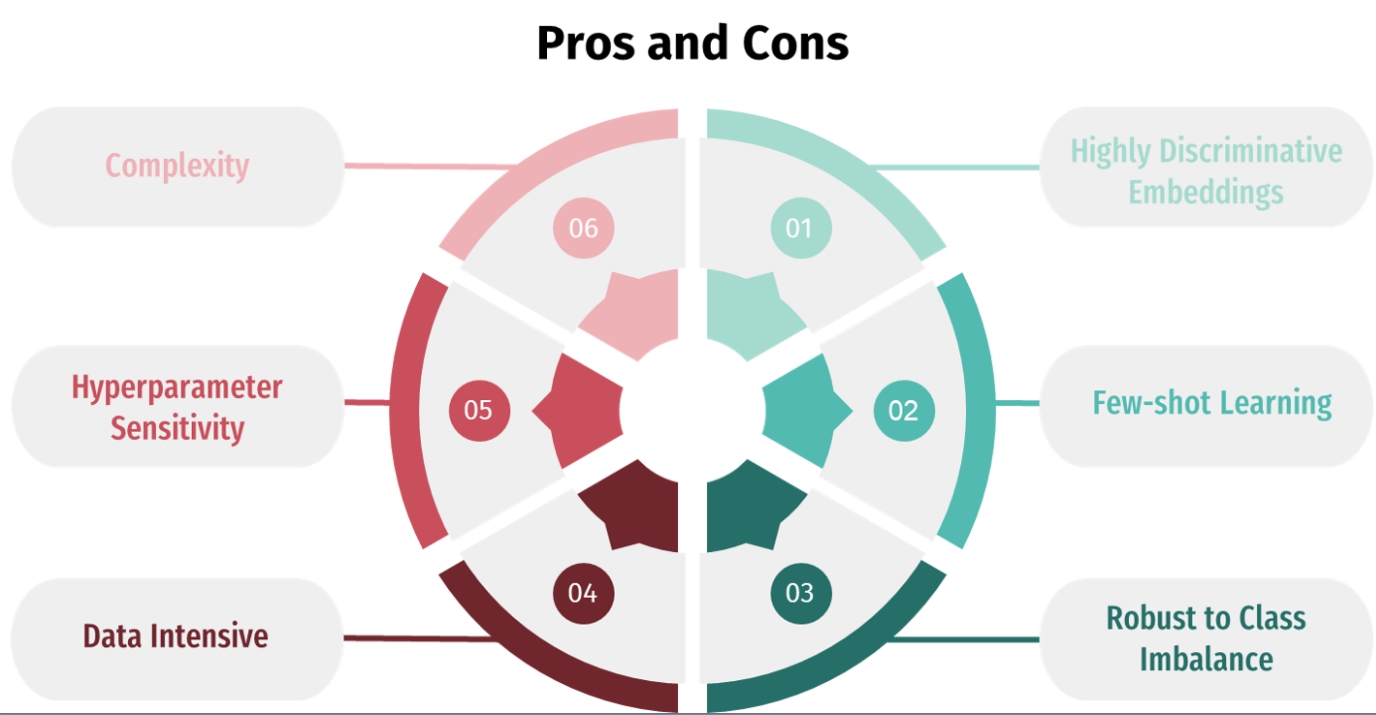}}
\end{center} 
\vspace{-0.7cm}
\caption{Advantages and challenges of implementing the Siamese network in a CBHIR. Points 1 to 3 mention the difficulties in using Siamese. Points 4 to 6 are about the positive points of using the Siamese network. }
\label{fig:pros}
\end{figure*}

The Siamese network is characterized by its twin neural network structure. Siamese networks are designed to compare the similarity between pairs of inputs. This network can be effective even with limited images in the train set, and it learns the representative features by focusing on the relative relationships between the pairs of input. Siamese networks bring advantages to the CBHIR approach, such as robustness to class imbalance and learning from semantic similarity. Figure \ref{fig:pros} illustrates the pros and cons of applying the Siamese network to histopathological images. Few-shot learning in the Siamese network empowers the model to be able to recognize patterns and measure the similarity between patches with very limited labeled data. Besides all the benefits of this network, it is important to mention that its performance might be sensitive to the hyperparameter settings and it requires extensive hyperparameter tuning. The Siamese network can be complex to design and train compared to the traditional image retrieval techniques. 

\begin{figure*}[htp!]
\begin{center}
\centerline{\includegraphics[width=0.88\textwidth]{ 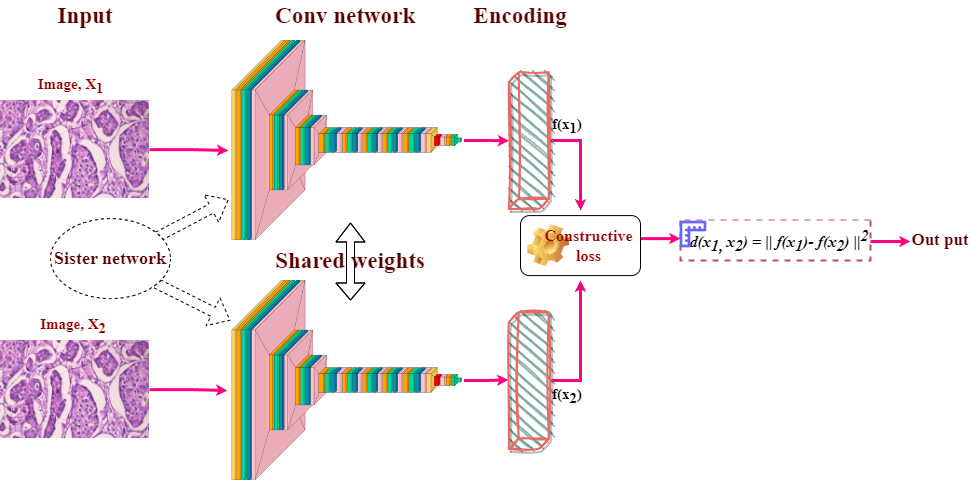}}
\end{center} 
\vspace{-0.7cm}
\caption{An overview of the proposed network. This is plotted with BreaKHis images. Colors of orange, red, pink, green, and teal represent Convolutional layers, MaxPooling2D, Dropout, and Flatten, respectively.}
\label{fig:overview}
\end{figure*}

Figure \ref{fig:overview}\footnote{The figure is plotted by breast histopathological images. The same pipeline was applied to the skin data set with a small difference in the internal structure of the Siamese network. The structure of these convolutional networks is slightly different for skin and breast data sets regarding filter size due to the difference between the size of images in each data set. More details are mentioned in section \ref{section:skin_details}} shows an overview of the proposed Siamese network on the BreaKHis data set. As can be seen, there are two identical networks, which are named sister networks. Each of these sisters contains the same architecture comprising a deep convolutional network, an encoding layer, and a distance metric. The main objective of these sisters is to minimize the contrastive loss in order to minimize the distance between images with the same histological features.

The Siamese network receives pairs of images as input. Each sister network takes one instance from the input pair and passes it through the convolutional layers of the proposed network for feature extraction. While training, these sisters shared the weights regarding encoding the input images to ensure that both networks learn to extract similar features of the input image. During training, in order to measure if each pair is similar or dissimilar, the label of the data is needed. Once the Siamese network learns to extract the representative features for pairs of input, it can be used to measure the similarity between pairs by computing the distance between the extracted features generated by the sister network for each patch instance.

\subsection{Contrastive loss}
The benefit of contrastive loss compared with the other loss functions is enabling the network to converge faster and reducing overfitting. This occurs since the network learns to be more generalized by focusing on the key features and determining the similarity.

The contrastive loss takes pairs of samples as anchor and neighbor or anchor and distant \cite{fischer2023self}. In the case of anchor and neighbor, they are pulled towards each other. In the other case, the distance between the anchor and the distant needs to be increased. The optimum embedding feature space is extracted while the anchor-distant distances get larger than the anchor-neighbor distances by a margin of $m$. This can enhance the embedding space and improve the discriminative capacity. The utilized contrastive loss defined for this work is as follows: 
\begin{equation}
    L_c = \sum_{i = 1}^b [(1-y) ||f(x_{1}^i) - f(x_{2}^i)||_{2}^2 + y[-||f(x_{1}^i) - f(x_{2}^i)||_{2}^2 + m ]_{+}]
\end{equation}\label{eq:loss}

\begin{figure*}[htp!]
\begin{center}
\centerline{\includegraphics[width=0.50\textwidth]{ 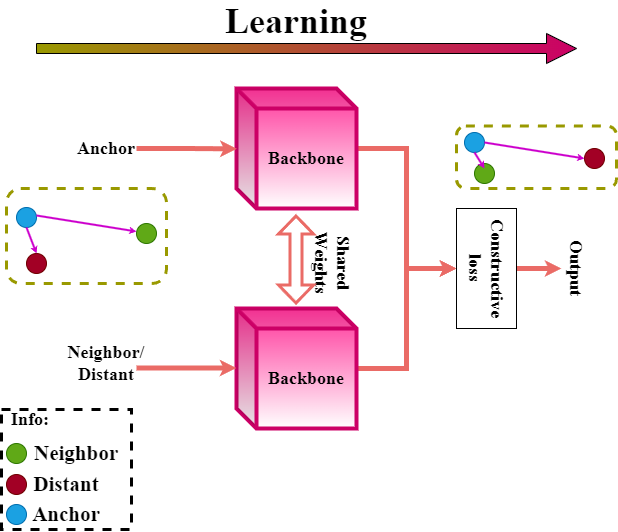}}
\end{center} 
\vspace{-0.7cm}
\caption{An overview of how contrastive loss receives the anchor-distant and the anchor-neighbor. The blue, red, and green points are anchor, distant, and neighbor samples.}
\label{fig:anchor}

\end{figure*}

The contrastive loss  ($L_c$) (1) should be minimized, which helps the network focus on capturing invariant features \cite{9206833}. $y$ is zero or one when the pair { $\{x_{1}^i, x_{2}^i\}$ } is anchor-neighbor and anchor-distant, respectively. We denote the $m$ as the margin and $f(x)$ as the output of the network (i.e., embedding). Let $b$ be the mini-batch size, and $||.||_{2}$ the $L_2$ norm \cite{1640964}. Figure \ref{fig:anchor} illustrates the workflow of the contrastive loss by receiving anchor-neighbor and anchor-distant. Three points in the colors red, green, and blue correspond to distant, neighbor, and anchor samples. Initially, the green point is located far from the anchor sample. As the model undergoes training, it acquires the ability to discriminate between samples. Ultimately, the contrastive loss function propels the red point to a distant location while maintaining the green point close to the anchor sample.

\subsection{Search engine}

\begin{figure*}[htp!]
\begin{center}
\centerline{\includegraphics[width=0.75\textwidth]{ 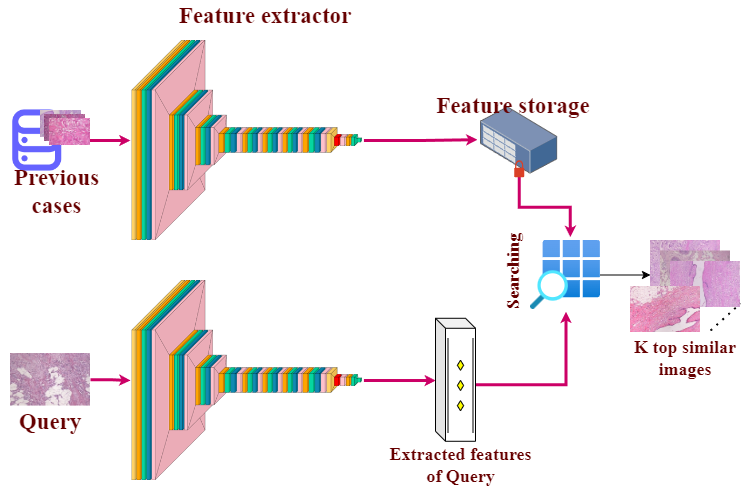}}
\end{center} 
\vspace{-0.7cm}
\caption{An overview of the proposed CBHIR by using the well-trained Siamese network as the FE.}
\label{fig:CBHIR_overview}
\end{figure*}

The trained Siamese network provides an FE that can extract the meaningful features of the data set. By reaching the well-trained FEs, it is time to employ them in a CBHIR framework. Figure \ref{fig:CBHIR_overview}\footnote{The figure shows the histopathological breast images. The same pipeline was applied for the skin images.} shows an overview of the proposed CBHIR. The trained Siamese network is used as an FE to extract two meaningful features of each input image. Each image $i$ in the data set is saved with a feature vector $F_i$ containing the two most representative features of that. All these feature vectors are saved in the feature storage.

When the CBHIR model receives a new query, the FE extracts the most meaningful features of the query ($F_Q$). As the next step, a distance function is needed to measure the distance of $F_Q$ with all saved $F_i$s in the feature storage ($F_D$). Among different types of distance functions, Euclidean distance was chosen for this paper as it is one of the most used metrics in order to calculate the distances between two images in CBHIR \cite{kumar2016adapting}. Euclidean distance is straightforward to understand geometrically and is computationally efficient. The other reason that the Euclidean function was selected is that it works well in high-dimensional vector spaces.

In the searching part of the retrieval step, the images are ranked with the shortest distance at the top, based on the Euclidean distance. This means the images with the most similarities are the top images. Then, top \textit{K} retrieved images are returned to pathologists for further investigations. So, it is expected to have the most similar images at the first top retrieved and the least similar image at the \textit{$K_{th}$} ($K>1$) retrieved. As much as the amount of \textit{K} is less, the number of images that are taken under evaluation is less. So, the model that reaches higher accuracy in less number of retrieving images is a more reliable tool. In this paper top, 1, 3, and 5 retrieved images on both data sets are shown.

\section{Material and Experimental setup}

\subsection{Material}
Two binary data sets of breast cancer and skin cancer were taken under study in this paper. Both data sets were stained with Hematoxylin and Eosin (H \& E) \cite{hoque2021retinex}. 
\subsubsection{Breast cancer data set}
BreaKHis data set contains $7909$ microscopic images of breast tumors at four magnification levels. This paper applied the experiments to images at $400\times$ magnification of this data set in the size of $224\times224\times3$ \cite{tabatabaei2023towards}. In grading breast cancer, the number of mitoses is a critical criterion. In order to measure it, $400\times$ magnification was selected, which is the highest level of magnification of BreaKHis. This binary data set comprises $1232$ malignant and $588$ benign at $400\times$ magnification \cite{tabatabaei2023wwfedcbmir}. In the training strategy, 70\% of the data set is randomly chosen for training, while the rest is considered for testing.

\subsubsection{Spitzoid melanocytic data set}

Spitzoid melanocytic is one of the most challenging skin tumors for pathologists because of its ambiguous histological features \cite{lodha2008discordance}. The data set was collected from 79 different patients containing 84 biopsies. Table \ref{tab:skin_data} describes the spitzoid melanocytic lesions database \cite{Laeti}. The data set is collected and annotated by the pathology laboratory from the University Clinic Hospital of Valencia. The data set contains uncommon skin tumors and $42$ STUMP images. 
\begin{table*}[htb!]
\centering
\caption{Description of the Skin data set. }
\label{tab:skin_data}
\begin{tabular}{ccc}
\hline
                      \textbf{Data type}&  \textbf{Malignant}& \textbf{Benign} \\ \hline
\multicolumn{1}{c|}{WSIs} & 34 & 50 \\
\multicolumn{1}{c|}{Patients} & 30 & 49 \\
\multicolumn{1}{c|}{Annotated} &  16 & 11 \\
\multicolumn{1}{c|}{Unannotated} & 29 & 41 \\ \hline
\end{tabular}
\end{table*}

Following \cite{Laeti}, the images were segmented from ROIs. The segmented images are non-overlapped, in the size of $512\times512\times3$. Following \cite{kanwal2023vision}, in our experiments, in order to prevent data leakage, the data set was later split into 70\% training and 30\% testing. In this paper, in the training, validation, and testing steps, the STUMP images were excluded. Later, in Section \ref{sec:STUMP}. they are considered a second test set, and the trained CBHIR is applied to them to retrieve the most similar patches to them. In this experiment, we can elucidate if the STUMP query provides more malignant or benign retrievals. The idea is to use the CBHIR as a transparent diagnostic tool for this kind of cases which often pose significant difficulties for pathologists.
\subsection{Experimental setting}
\begin{figure*}
\begin{center}
\centerline{\includegraphics[width=0.98\textwidth]{ 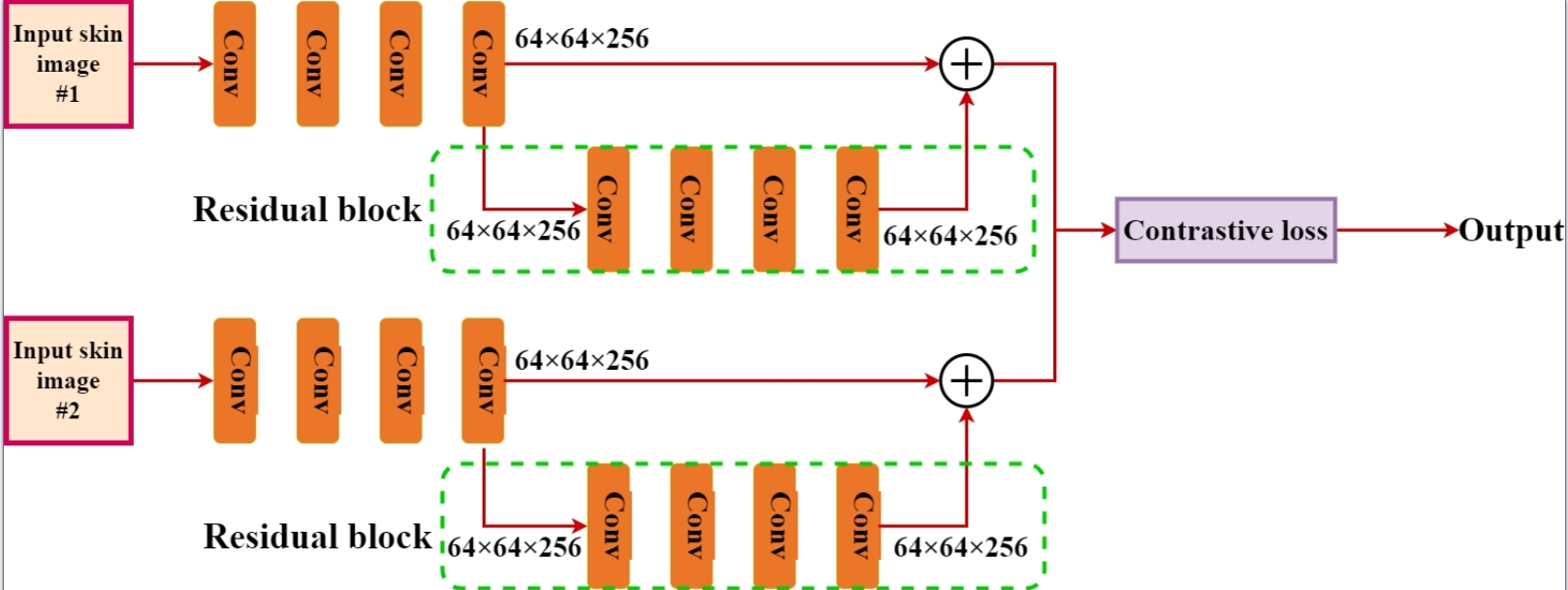}}
\caption{ An overview of the proposed method on skin images in perspective of convolutional layers. Each of the Convolutional layers in the figure is named \textit{Conv}.}
\label{fig:convs}
\end{center} 
\end{figure*}

The proposed convolutional network in this sister network comprises two stages of convolutional layers. More details about the Siamese structure and the hyperparameters for the breast (Breast-twins) and skin (Skin-twins) data set are mentioned in the following subsections and Figure \ref{fig:convs}. 

Tuning hyperparameters in Siamese networks is challenging for different purposes, and it relates to the type of the data set. So, it is important to note that Breast-twins and Skin-twins models were trained with a learning rate schedule from Keras with an initial learning rate of $0.01$, decay steps of $10000$, and decay rate of $0.9$. Stochastic Gradient Descent (SGD) optimized the contrastive loss while training. For both data sets, many trials and errors found the amount of margin ($m$), and $0.9$ was found as the optimum amount for this task.

First, the network is initialized with a contrastive loss and SGD as the optimizer. Then, the first image of the image pairs is processed through the network. Subsequently, the second image of the image pair is fed into the network. Then, the loss is determined by the outputs from the first and the second images. To optimize the model, the gradient is computed, and the model weights are updated using the Stochastic Gradient Descent (SGD) algorithm.

The experiments in this paper were run on GPU with the \textit{NVIDIA GeForce RTX 3090}.

\subsubsection{Breast-twins in details} \label{section:breast_details}
Breast-twins start with four convolutional layers ($[32,64, 128, 256]$) with the filter size of 3 by 3 while stride is $[1, 2, 2, 2]$. In order to extract deeper features, the network goes deeper with more convolutional layers with the stride of $1$ and a kernel size of 3 by 3. The number of convolutional filters in each layer of the residual block is $[64, 32, 1, 256]$ to reduce the spatial dimensions while increasing the number of filters and compressing the information. In this case, decreasing the filter size helped control the number of parameters while still capturing relevant features. Ultimately, a convolutional layer and a Global Max Pooling2D (GMP) are applied to reduce the spatial dimensions and retain important features.

In this experiment, the Breast-twins network was trained over 300 epochs with a batch size of 16.
\subsubsection{Skin-twins in details}
\label{section:skin_details}
In the case of the skin data set, since the size of the patches is bigger than BreaKHis, different filter sizes are needed. So, as can be seen in Figure \ref{fig:convs}, first, four convolutional layers with the filter size of 3 by 3 and the layers of $[32, 64, 128, 256]$ and stride of $[1, 2, 2, 2]$ compress the input. In this part of the network, we aim to gradually reduce the spatial dimensions by increasing the filter sizes. Then, in order to get more meaningful features of the input, we added a new series of convolutional layers of [$32, 16, 1, 256]$ with stride fixed to $1$ and filter size of $3 \times 3$ in the residual block of the proposed architecture. This makes the network deep enough but not excessively complex. The Skin-twins network was trained over 300 epochs with a batch size of 16. Figure \ref{fig:convs} provides an overview of the proposed structure from the convolutional point of view. 

In the training part of the Skin-twins, the skin data set is considered a binary data set and excludes the STUMP class of the data set to use as the test set for the experiments in section \ref{sec:STUMP}.

\section{Results and Discussion}
This section reports the results of Breast-twins and Skin-twins with tables of the results and visual evaluations to provide a comprehensive comparison.

\subsection{Results of Breast-twins}

Table \ref{tab:break} provides a comparison between some state-of-the-art papers that reported the results with different amounts of \textit{K}. In this table, the amount of \textit{K} for each study is mentioned since to the best of the author's knowledge, there are not any papers reporting the results at the first top retrieval on a breast data set. As can be seen in Table \ref{tab:break}, the Breast-twins network is able to correctly retrieve the first top image with a $59\%$ of accuracy. 

\begin{table}[htp!]
\centering
\caption{ Comparison between the cutting-edge methods and Breast-twins model. This comparison is on BreaKHis at $400\times$ magnification. }
\begin{tabular}{|c|c|c|c|c|}
\hline
 \textbf{Method}&\textbf{K}  &\textbf{Accuracy}  & \textbf{Precision} &  \textbf{F1score}\\ \hline\hline
 \textbf{Breast-twins}& 1 & 0.59 & 0.67 & 0.70 \\ 
  \textbf{Breast-twins}& 3 & 0.83 & 0.81 & 0.88 \\ 
  \textbf{Breast-twins}& 5 & 0.92 & 0.90 & \textbf{0.94} \\ 
  \textbf{MCCH} \cite{minarno2021cnn}& 5 & - & 0.89 & -\\ 
  \textbf{FedCBMIR} \cite{tabatabaei2023wwfedcbmir} & 5 & 0.96 &\textbf{0.94} & - \\
  \textbf{HSDH} \cite{alizade}& 400 & \textbf{0.99} & - & - \\ \hline
 
\end{tabular}\label{tab:break}
\end{table}

Paying attention to the results, it can be understood that with $\textit{K} = 5$, Breast-twins outperforms MCCH \cite{minarno2021cnn}. Breast-twins models has only a $0.06$ difference in terms of accuracy with HSDH \cite{alizade}, which sets the amount of $\textit{K} = 400$, which is a high amount for a binary data set. On the other hand, although FedCBMIR in \cite{tabatabaei2023wwfedcbmir} was trained on all four magnifications and then tested on BreaKHis at $400\times$, it has only $0.04$ higher accuracy than the Breast-twins model which only trained on BreaKHis at $400\times$ magnification. According to Table \ref{tab:break}, the performance of Breast-twins at top $1, 3,$ and $5$ shows its reliability.

Table \ref{tab:breast_compare_K} compares the performance metrics, including accuracy, recall, precision, and F1score, between the proposed Breast-twins model with the previously proposed CAE \cite{tabatabaei2023towards}. The results showcased in this table demonstrate that although the CAE could provide a comparable performance in terms of accuracy, precision, and F1score at the top 3, there is a notable disparity in its performance regarding the first top retrieval. This table can clarify the potential risks of a CBHIR technique with high accuracy only at top \textit{K} while $K > 1$.

\begin{figure*}[hpt!]
\begin{center}
    \includegraphics[width=.3\textwidth]{ 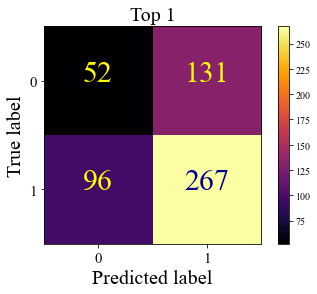}
    \includegraphics[width=.3\textwidth]{ 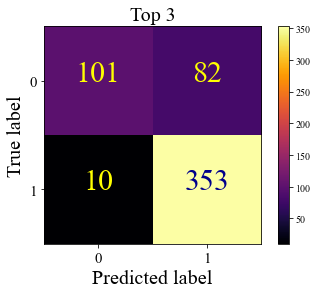}
    \includegraphics[width=.3\textwidth]{ 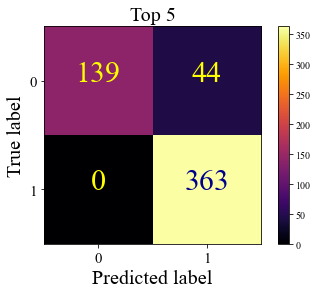}
    \caption{Three CMs at top $1, 3$, and $5$ for Breast-twins CBHIR model.}\label{fig:CM_breast}
\end{center}

\end{figure*}

\begin{table}[htp!]
\centering
\caption{A comparison between the CAE and Breast-twins at \textit{K}= 1 and 3.}
\label{tab:breast_compare_K}
\begin{tabular}{|c|c|c|c|c|c|} \hline
 \textbf{Method}& \textbf{K} & \textbf{Accuracy} &  \textbf{Recall} &\textbf{Precision} & \textbf{F1score}  \\ \hline\hline
 \textbf{CAE} & 1 & 0.48 & 0.54 & 0.64 & 0.56 \\
 \textbf{Breast-twins} & 1 & \textbf{0.58} & \textbf{0.73} &\textbf{0.67} &\textbf{0.70} \\ \hline\hline
 \textbf{CAE} & 3 & 0.83& 0.95&\textbf{0.82} &0.88 \\
 \textbf{Breast-twins} & 3 & \textbf{0.83} &\textbf{ 0.97} & 0.81& \textbf{0.88}\\
 \hline
\end{tabular}
\end{table}
Finding a sufficient amount of \textit{K} is challenging for DL experts. For instance, in the case of \cite{alizade}, the results were reported at the top 400, which delivered $99\%$ accuracy for a binary data set. According to the definition of accuracy in CBHIR, increasing the amount of \textit{K} yields higher accuracy. But a high amount of \textit{K}, means retrieving images with longer distances and less similarity. Therefore, a CBHIR with high accuracy at a high amount of \textit{K} cannot provide a confident tool. The impacts of the amount of \textit{K} can be seen in Figure \ref{fig:CM_breast} which shows three Confusion Matrix (CM) at top 1, 3, and 5 for breast cancer.

\begin{figure*}[htp!]
    \centering
    \begin{minipage}{0.48\textwidth}
        \includegraphics[width=\linewidth]{ 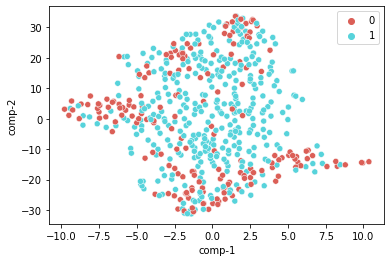}
        \subcaption{Embedding space extracted by FedCBMIR}
    \end{minipage}\hfill
    \begin{minipage}{0.48\textwidth}
        \includegraphics[width=\linewidth]{ 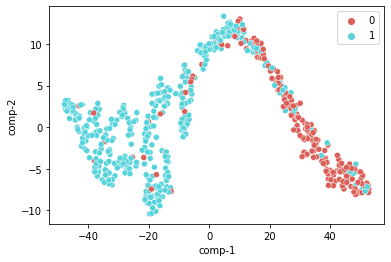}
        \subcaption{Embedding space extracted by Breast-twins}
    \end{minipage}
    
    \caption{TSNE plot of the extracted feature space by applying two different FE to the breast data set. \textbf{Zero} label in red shows the embedding space of benign tissues and the \textbf{One} labeled points in blue correspond to the malignant cases.}
    \label{fig:TSNE_breast}
\end{figure*}

Figure \ref{fig:TSNE_breast} plots the T-distributed Stochastic Neighbor Embedding (TSNE) of the extracted features of the breast data set. TSNE is a data visualization technique that is used for reducing the dimensionality of high-dimensional data while preserving the similarities between data points. Figure \ref{fig:TSNE_breast} compares the extracted embedding space as a result of FedCAE \cite{tabatabaei2023wwfedcbmir} and the proposed Breast-twins. FedCAE was trained on all four magnifications of the BreaKHis data set and then applied to BreaKHis at $400\times$ magnification. As can be seen in Figure \ref{fig:TSNE_breast}, although FedCAE is more generalized due to the different magnifications, the embedding space resulting from Breast-twins is more discriminative.

In breast cancer diagnosis, pathologists need to take into account the histopathological patterns of the tissue to be able to grade the tissue. Well-differentiated patterns, low nuclear pleomorphism, and mitotic activities are some of the important criteria that pathologists analyze in their query to write a diagnosis report. These grading patterns assist pathologists in determining the best treatment and predicting the likelihood of the tumor spreading. So, the CBHIR system answers these demands from pathologists by providing similar patterns for the pathologists.

\begin{figure*}[htp!]
\begin{center}
\centerline{\includegraphics[width=0.93\textwidth]{ 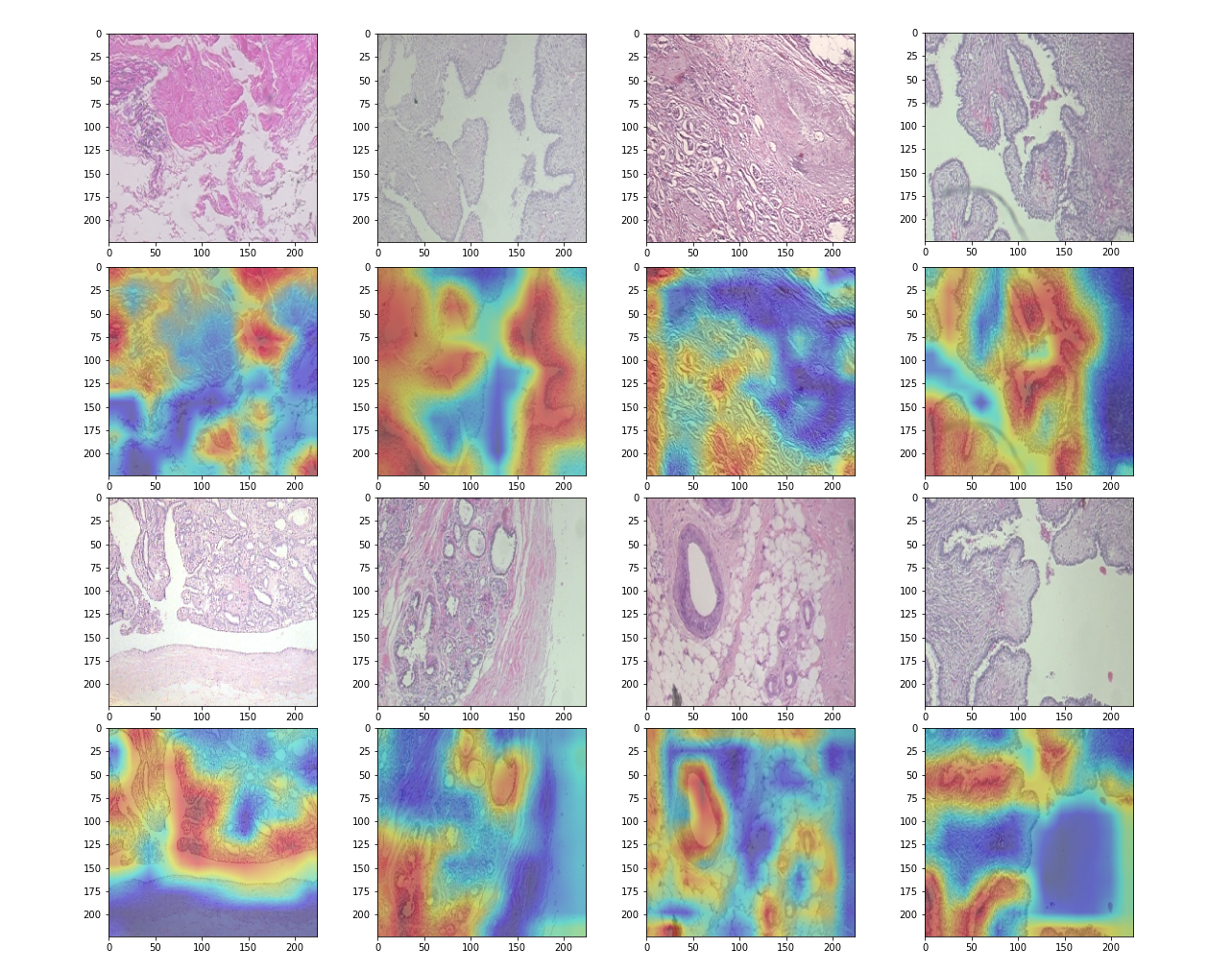}}
\end{center} 
\vspace{-0.7cm}
\caption{ Eight random patches of breast cancer data set. The odd rows show the original images. The even rows are the Grad-CAMs mapped on top of the original images. }
\label{fig:gradcam_breast}
\end{figure*}

Figure \ref{fig:gradcam_breast} shows eight random patches of BreaKHis data set with their Grad-CAM plots \cite{zhang2022histokt}. The presented Grad-CAM visualizations were generated by fusing the GMP layer of the proposed Breast-twins network to emphasize the final feature representation of the model. 

These Grad-CAM plots provide insight into the DL-based Siamese networks to clarify why the model made a particular decision. This means that it can identify if the model focused on the correct histopathological features or retrieved the images based on the artifact. Figure \ref{fig:gradcam_breast} clarifies the important regions of the tissue that the Breast-twins network detected as an important part of the patches for the final decision of the search engine in the CBHIR model. This figure proves the interpretability of the proposed model.

\begin{figure*}[hpt!]
\begin{center}
\centerline{\includegraphics[width=0.95\textwidth]{ 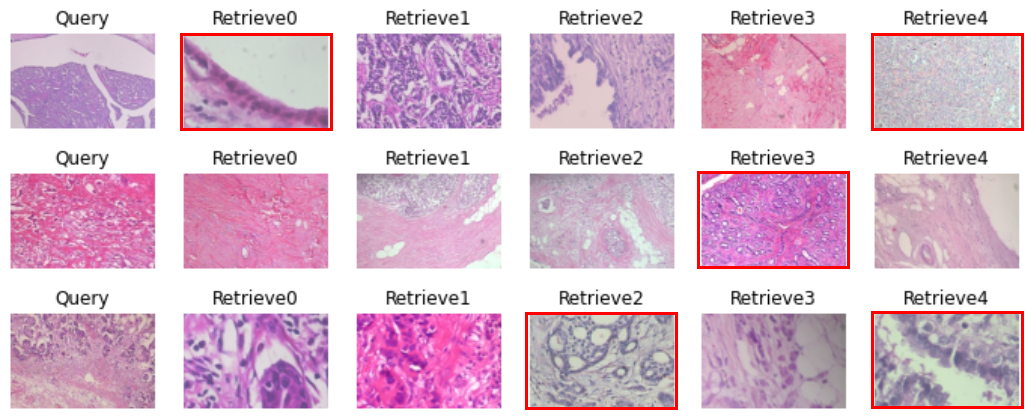}}
\end{center} 
\vspace{-0.7cm}
\caption{ Three random breast queries with their 5 top retrieved images. The miss-retrieved images are marked in red.}
\label{fig:return_breast}
\end{figure*}

Figure \ref{fig:return_breast} visualizes the output of the proposed CBHIR by using Breast-twins for three random queries. The retrieved images were compared with the query based on their labels. Images with a different label than the query are marked in red. It is seen that the proposed Breast-twins network has the ability to return the patches with similar histopathological features with the correct class labels of these patches. Results indicate that the Breast-twins network is highly efficient in retrieving similar patches.

\subsection{Results of Skin-twins }
Table \ref{tab:skin} illustrates the performance of Skin-twins at top 3 and 5. Skin-twins can find similar patches to the query from the spitzoid cancer data set, accurately with $64\%$ and $94\%$ of F1score for the classes of \textit{Benign} and \textit{Malignant}, respectively. 

\begin{table}[htp!]
\centering
\caption{Results of Skin-twins in searching between "\textit{Malignant}" and "\textit{Benign}" at top 3 and 5. \textit{K} = 3, 5) }
\label{tab:skin}
\begin{tabular}{|c|c|c|c|cc|c|c|}
\hline
  \textbf{Method}& \textbf{ \textit{K}}& \textbf{Precision} & \textbf{Recall} & \multicolumn{2}{c|}{\textbf{F1score}} & \textbf{F1s-avg} &\textbf{Accuracy}  \\
 \multirow{2}{*}{\textbf{Skin-twins}}& \textbf{3} & 0.88 & 0.99 & 0.64 & 0.94 & 0.94 & 0.89\\
 &\textbf{5} &  0.92& 1 & 0.80 & 0.96& 0.96& 0.93\\
\hline
\end{tabular}
\end{table}
Table \ref{tab:skin_comparision} compares the proposed model with our previously proposed CBMIR model based on CAE \cite{tabatabaei2023towards} at the top first retrieved images. As can be understood from the reported results, Skin-twins framework reaches $67\%$ higher precision at the top first retrieved images, which is high development in CBHIR systems.

\begin{table}[htp!]
\centering
\caption{A comparison between the results of Skin-twins and state-of-the-art studies at the first top retrieval \textit{K} = 1).}
\label{tab:skin_comparision}
\begin{tabular}{|c|c|c|c|c|}
\hline
 \textbf{Method} & \textbf{Precision} & \textbf{Recall}  &  \textbf{F1score} & \textbf{Accuracy} \\ 
 \textbf{CAE} & 0.13 & 0.14 &0.135 & \textbf{0.72}  \\
 \textbf{Skin-twins} & \textbf{0.80} & \textbf{0.82} & \textbf{0.81} & 0.69\\ \hline
\end{tabular}
\end{table}

\begin{figure*}[hpt!]
\begin{center}
    \includegraphics[width=.3\textwidth]{ 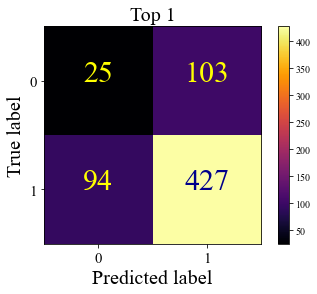}
    \includegraphics[width=.3\textwidth]{ 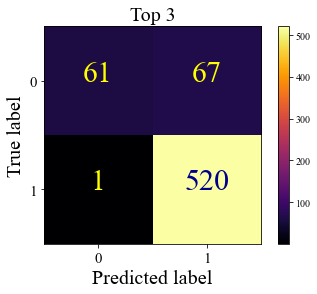}
    \includegraphics[width=.3\textwidth]{ 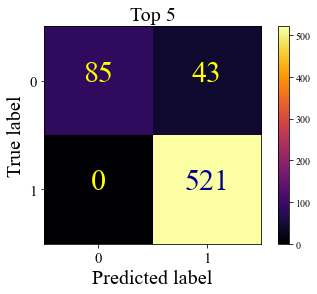}
    \caption{Three CMs at top $1, 3$, and $5$ for Skin-twins CBHIR model.}\label{fig:CM_skin}
\end{center}

\end{figure*}

Figure \ref{fig:CM_skin} shows the CM of the proposed method on the skin data set at top  \textit{K} = $1, 3$, and $5$. In these CMs, \textit{Benign} and \textit{Malignant} are labeled as zero and one. The figure illustrates that by increasing the amount of \textit{K}, the amounts of True positive and True negative are increasing.
\begin{figure*}[htp!]
\begin{center}
\centerline{\includegraphics[width=0.55\textwidth]{ 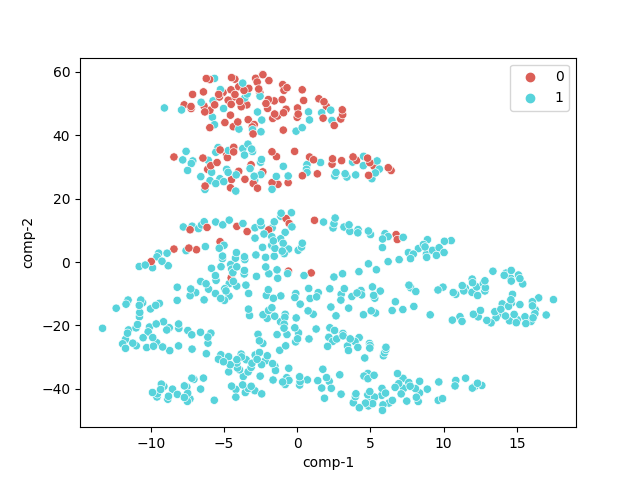}}
\end{center} 
\vspace{-0.77cm}
\caption{ TSNE plot of the extracted features of skin data set by Skin-twins network. Features corresponding to benign tissues are plotted in red and labeled as \textbf{Zero}. The blue points are related to the malignant tissues which are labeled as \textbf{One}. }
\label{fig:TSNE_skin}
\end{figure*}
Figure \ref{fig:TSNE_skin} shows the TSNE plot of the extracted features of the skin data set. Each point in this plot corresponds to a data point, and the proximity of points indicates similarity in the underlying feature space. The points are color-coded to represent the benign and malignant classes in red and blue, respectively. This figure provides insight information of the inherent patterns and relationships of the embedding space.

\begin{figure*}[htp!]
\begin{center}
\centerline{\includegraphics[width=0.67\textwidth]{ 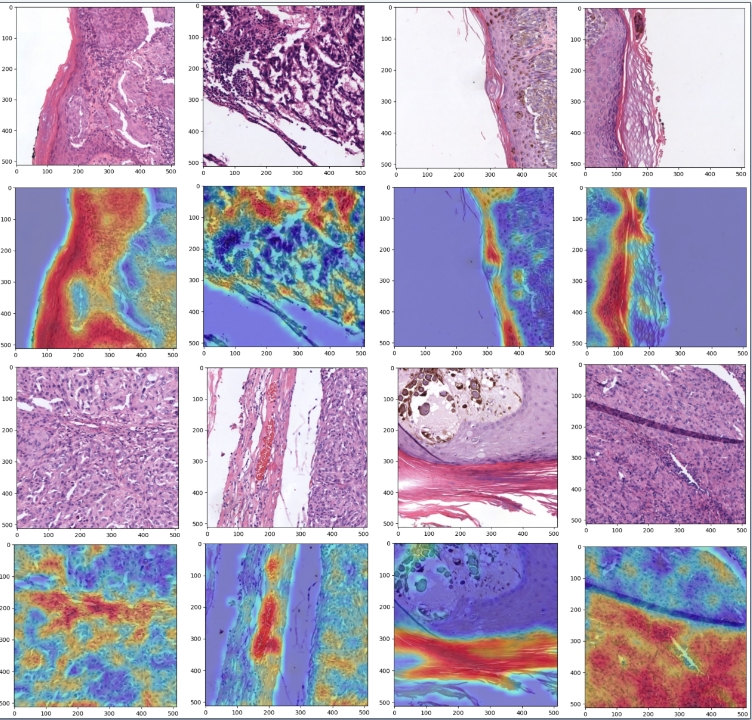}}
\end{center} 
\vspace{-0.7cm}
\caption{ Eight random patches of skin cancer data set. The odd rows show the original images. The even rows are the Grad-CAMs mapped on top of the original images. }
\label{fig:gradcam_skin}
\end{figure*}

The Grad-CAM plots of the Skin-twins are shown in Figure \ref{fig:gradcam_skin}. In skin cancer detection, Cellularity, Nuclear Features, Cellular Morphology, Architecture, Mitotic activity, Stroma and Connective Tissue, etc., are some of the histological features that pathologists consider to grade a skin cancer \cite{mosquera2022deep}. These are the histological patterns that the CBHIR should focus on to find similar patches. Figure \ref{fig:gradcam_skin} shows the success of the model in detecting the cancerous and abnormal regions of the tissue. For instance, in some cases, the main focus is cellularity and assessing the density of the cells within the tissue. High cellularity may indicate increased cell proliferation or inflammation. So, the Grad-CAM plots generate a visually interpretable explanation for the retrieved images, which explains why the Skin-twins model retrieved the images for the pathologists.

\begin{figure*}[htp!]
\begin{center}
\centerline{\includegraphics[width=0.70\textwidth]{ 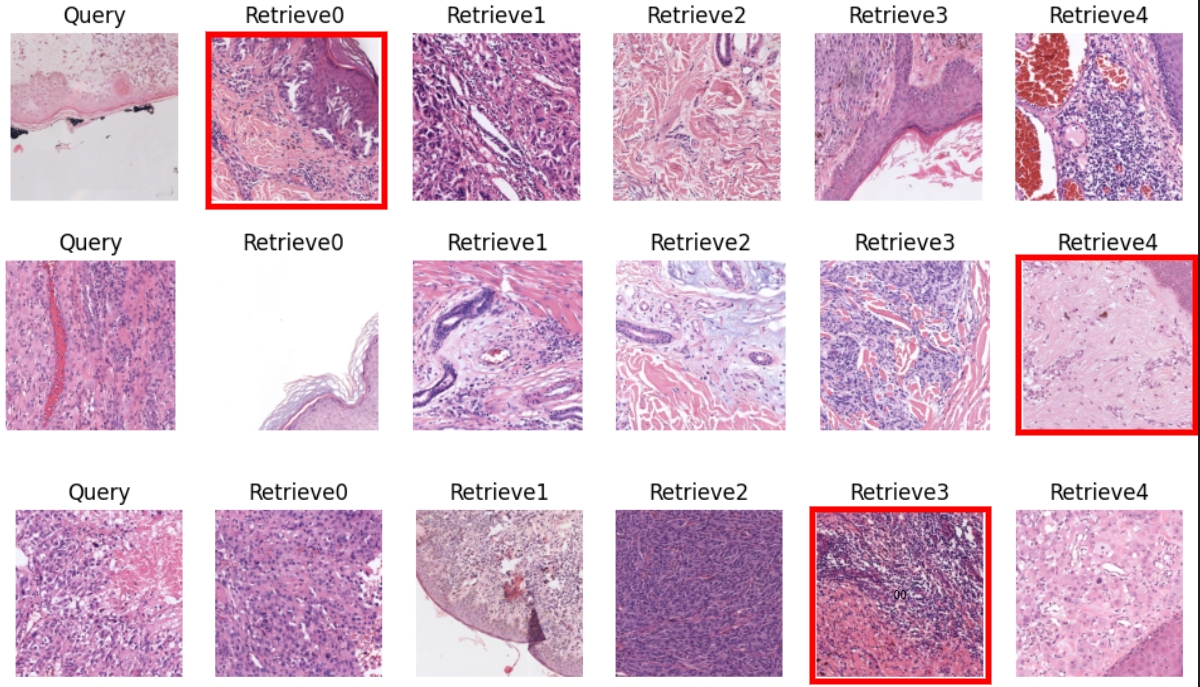}}
\end{center}
\vspace{-0.7cm}
\caption{ Three random images with their top 5 retrieved images. The returned images with a different label are marked in red. }
\label{fig:return_skin}
\end{figure*}

Figure \ref{fig:return_skin} depicts three query examples of the skin data set with their retrieved images. The red line around the retrieved patches clarifies that the retrieved image does not have the same label as the query. As can be seen in this figure, just one out of five retrieved patches was not at the same cancer grade as the query.

\subsection{STUMP searching}\label{sec:STUMP}
\begin{figure*}[htp!]
\begin{center}
\centerline{\includegraphics[width=0.85\textwidth]{ 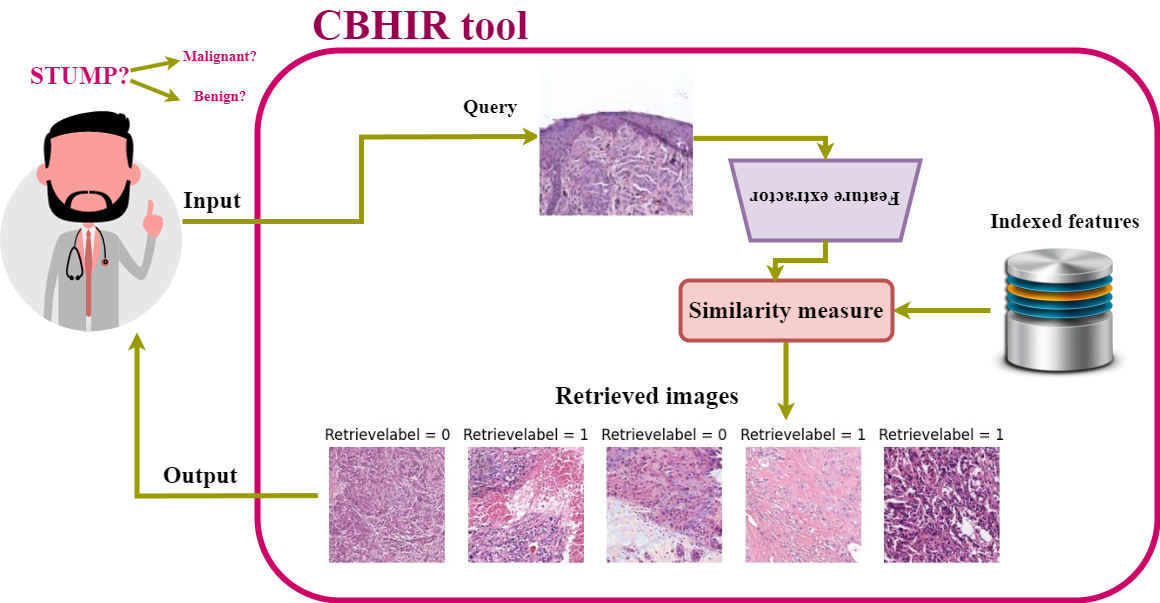}}
\end{center} 
\vspace{-0.7cm}
\caption{ A real-world scenario that a pathologist faces with a STUMP case and uploads the query as input to the proposed CBHIR approach. First, the model feeds the query to the proposed FE and extracts the query features. Second, a similarity measure applies to the query features and indexed features of the database. Then, it is time to rank and visualize the top 5 similar patches to pathologists as the output of the proposed CBHIR. }
\label{fig:stum}
\end{figure*}

Spitzoid lesions present a diagnostic conundrum due to their intricate histology, creating challenges in establishing clear parameters between benign nevi and potentially malignant melanomas. In particular, STUMP cases require careful assessment to determine their true nature. In this paper, the model was isolated from STUMP images while training and testing steps for the above experiments. Figure \ref{fig:stum} explains that this experiment mimics the situation that pathologists face with a STUMP tissue and need a second opinion to assist them in writing the diagnosis report. To do so, the Skin-twins model, which is trained on the binary data set including \textit{Benign} and \textit{Malignant}, is fed by the STUMP queries. The Skin-twins model displays top \textit{K} similar patches to pathologists. The top \textit{K} return patches with their labels providing some clues and hints regarding histopathological features and patterns for the pathologists. 

\begin{figure*}
\begin{center}
\centerline{\includegraphics[width=1.1\textwidth]{ 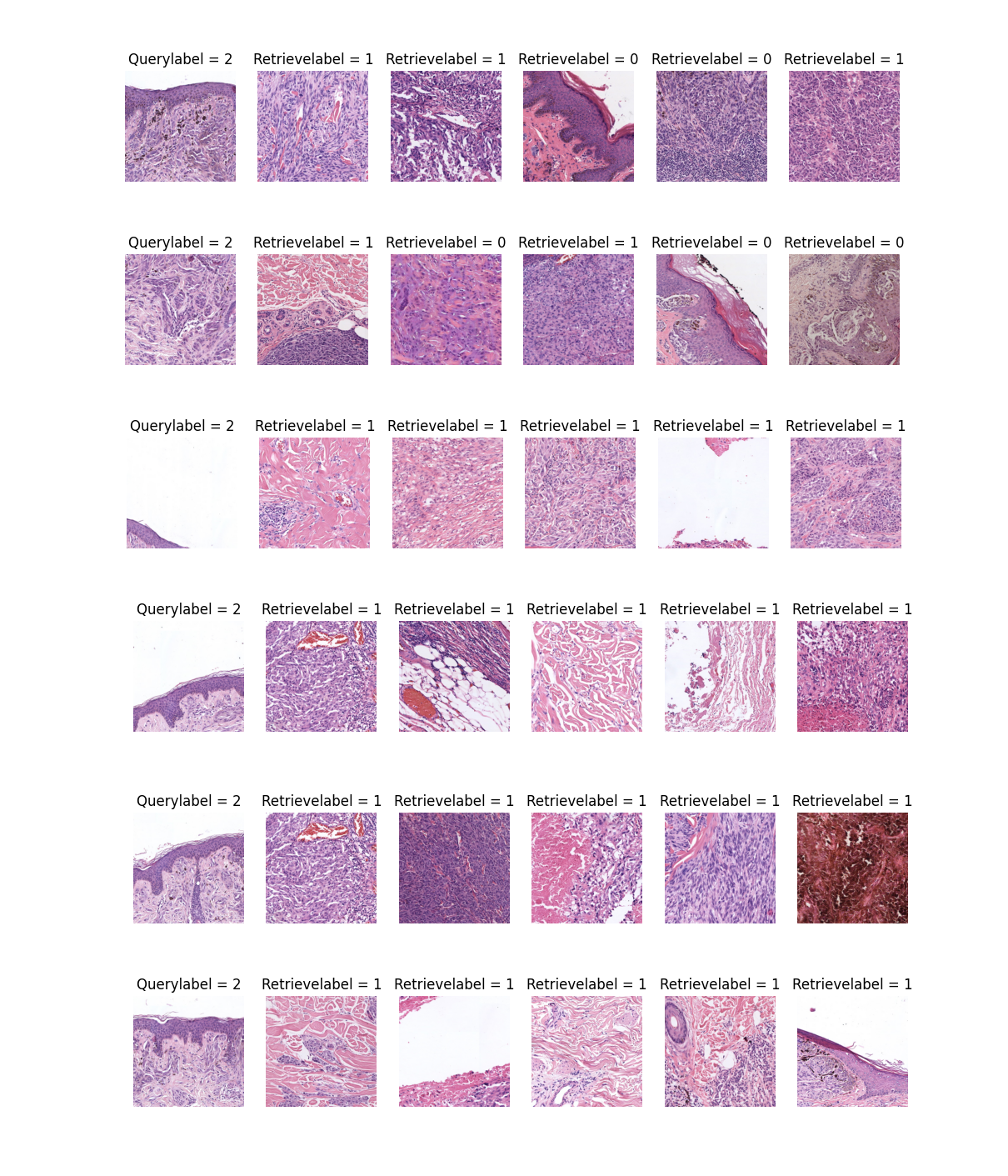}}
\vspace{-0.7cm}
\caption{ Six random examples of STUMP queries and their retrieved images. The STUMP queries were labeled = $2$, Malignant = $1$, and  Benign = $0$. }
\label{fig:STUMP}
\end{center} 
\end{figure*}

In Figure \ref{fig:STUMP}, for each query STUMP (labeled $= 2$), five top similar patches and their labels are retrieved from the data set, including \textit{Benign} (labeled $= 0$) and \textit{Malignant} (labeled $= 1$). In this case, pathologists can analyze the patterns in the retrieved tissues and compare them with the histopathological features of their STUMP query. The final decision depends on the pathologists' point of view and their knowledge. For instance, in the case of line 3 in Figure \ref{fig:STUMP}, it can be concluded that the STUMP tissue is Malignant since all the retrieved images are labeled $ = 1$.

\begin{figure*}[htp!]
\begin{center}
\centerline{\includegraphics[width=0.99\textwidth]{ 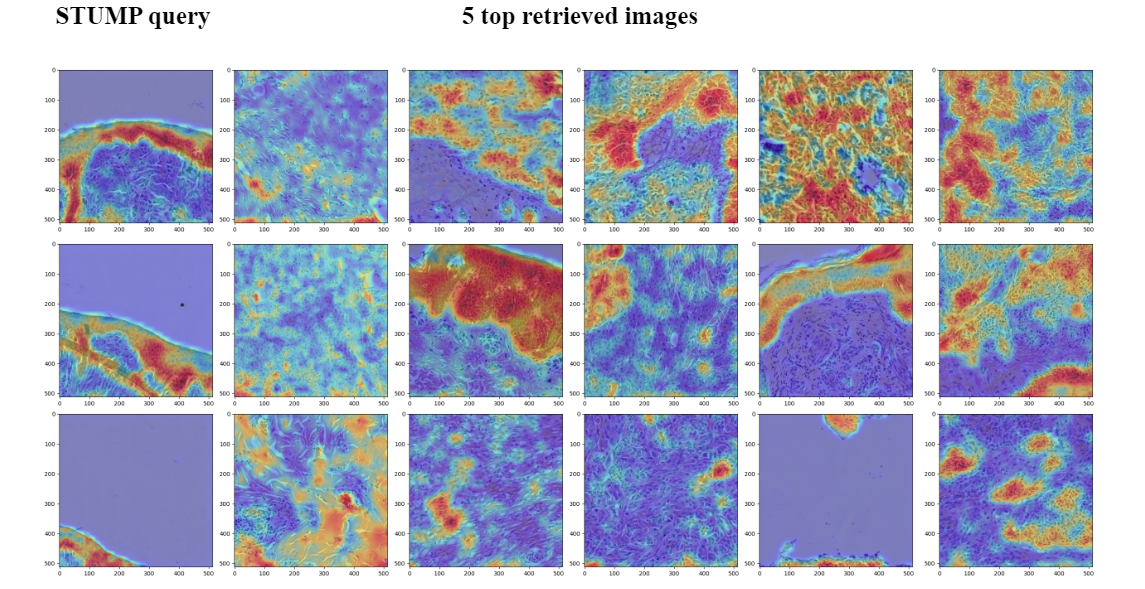}}
\end{center} 
\vspace{-0.7cm}
\caption{ Three random STUMP patches with their top 5 retrieved image. The Grad-CAMs show the main patterns that the CBHIR took into account to find the similarity between the query and the retrieved patches. }
\label{fig:stum_return}
\end{figure*}

In Figure \ref{fig:stum_return} there are Grad-CAM plots of three random STUMP queries with their top 5 retrieved images. The histopathological patterns and the regions of interest at the patch level that the Skin-twins network paid attention to retrieve the patches related to the STUMP queries are highlighted. As can be seen in Figure \ref{fig:STUMP} and Figure \ref{fig:stum_return}, pathologists can diagnose and grade the tissue more confidently by receiving the top 5 similar patches and their corresponding grades in addition to the highlighted histopathological patterns and the tumor cells.

\subsection{Comparison with classification approaches}
Classification delivers the label of the input image to provide a second opinion on the query tissue for the pathologists. While, CBHIR not only provides the label, but also retrieves similar patches. The outputs of CBHIR assist pathologists in finding the relevant histopathological features in the return images. 

The only similarity between a CBHIR and a classifier is that both report the labels. A classifier has done this as a black box, while in CBHIR, pathologists have this opportunity to analyze the retrieved patches by their knowledge to understand why the model retrieves images. So, a classification framework by its nature, directly reports the label. However, in CBHIR the label is obtained based on the label of the retrievals and the final decision of the pathologists by analyzing the patterns of the query and the returned images from a CBHIR model.

\begin{landscape}
{
In order to evaluate the performance of the CBHIR, the retrieved images have to be in the same cancer type as the query. Table \ref{tab:classifier_skin} presents a comparative analysis between the Skin-twins CBHIR model and the classifiers reported in \cite{Laeti} which have been recently published.

In this table, the results of Skin-twins are reported in the top 3 retrieved images. The obtained results of Skin-twins CBHIR could exhibit higher performance compared with all the mentioned methods in terms of F1score. Furthermore, assume that \textit{K} was set as $5$ as the most used amount of \textit{K} in the previous studies \cite{tabatabaei2023wwfedcbmir, Zahra_IVMSP, alizade, zheng2017size, SMILY, tabatabaei2023towards, kalra2020yottixel}, the F1score and accuracy are $96\%$ and $93\%$, respectively, much higher than the results of classifiers. 

Table \ref{tab:classifier_breast} provides the comparison between the Breast-twins CBHIR model at the top 5 and the recent classifiers on the BreaKHis data set. Breast-twins CBHIR could reach a higher accuracy than the recent classifiers, although, in some studies, it is slightly under-performing a classifier. 

\begin{table*}[htp!]
\centering
\caption{A comparison between classifiers and the Skin-twins CBHIR, $K = 3$.}
\label{tab:classifier_skin}
\begin{tabular}{|c|c|c|c|c|c|}
\hline
 \textbf{Method}& \textbf{VGG16} \cite{Laeti} & \textbf{ResNet-34} \cite{Laeti} & \textbf{ResNet-50} \cite{Laeti} &\textbf{ResNet-34 (\textit{$\overset{\alpha}{S}$ })} \cite{Laeti} & \textbf{Skin-twins} \\ 
 \textbf{F1score}& 0.70  & 0.84 & 0.76  & 0.74 &  \textbf{0.94} \\ 
 \textbf{Accuracy}& 0.72 & 0.85  & 0.77  & 0.68  & \textbf{0.90} \\ \hline
\end{tabular}
\end{table*}

\begin{table}[htp!]
\caption{A comparison between the recently proposed classifiers and the Breast-twins CBHIR, $K = 5$. }
\centering
\label{tab:classifier_breast}
\begin{tabular}{|c|c|c|c|c|c|c|}
\hline
\textbf{Method}  & \textbf{\shortstack{PFTAS\\ + SVM}} \cite{spanhol2015dataset}  & \textbf{IDSNet} \cite{li2020classification}& \textbf{\shortstack{FE\\-VGGNET16\\-SVM(POLY)}} \cite{kumar2020deep} & \textbf{\shortstack{FCN-Bi\\-LSTM}} \cite{budak2019computer} & \textbf{\shortstack{AE + \\Siamese \\Network}} \cite{liu2022breast}& \textbf{\shortstack{Breast\\-twins}} \\
\textbf{Accuracy} &    0.823&  0.845         &  0.934         &   0.942        &    \textbf{0.967 }      &       0.92    \\ \hline
\end{tabular}
\end{table}

} \end{landscape}

For instance, in \cite{liu2022breast}, the authors trained the model twice. Once the model was trained as an Auto Encoder (AE), the model was re-trained as a Siamese classifier. This double-trained AE + Siamese network could deliver $0.04$ amount more accuracy than the proposed Breast-twins CBHIR model. These two tables (Table \ref{tab:classifier_breast} and Table \ref{tab:classifier_skin}) prove that the proposed CBHIR models could only retrieve images within the same cancer type thanks to the Breast/Skin-twins.

\section{Conclusions and future lines}
In this paper, we have proposed a novel Content-Based Histopathological Image Retrieval (CBHIR) approach on breast and skin cancer data sets based on a costume-built Siamese network. These methods are named Breast-twins and Skin-twins in this paper since it is a pairwise framework. In order to train them, a contrastive loss function, which is a distance-based loss, has been applied.

The proposed Siamese network is used to extract the discriminative features of the patches to feed to the search engine. In the retrieval step, the trained model compares the query image with all the images in the data set to find and rank similar patches based on Euclidean distance. In this paper, we evaluated the performance of the approach on both data sets at the top 1, 3, and 5. To the best of the author's knowledge, this is the first time that a CBHIR approach reached a high performance at the top first retrieved patches on breast and skin cancer. This benefit of the proposed approach makes it more reliable for pathologists to consider it in their daily workflow.

On one hand, the proposed Breast-twins network works with the highest magnification available in the BreaKHis data set. It is designed to work on images at $400\times$ magnification since it can assist pathologists in measuring the amount of mitosis as an important criterion in breast cancer grading. We compared the performance of the Breast-twins network with some state-of-the-art CBHIR techniques. As a result of the comparisons, the proposed CBHIR could overpass the other techniques with a significant difference at the top first retrieved patches. Moreover, by comparing the obtained results with cutting-edge classifiers, we illustrated the high performance of the model in retrieving the patches in the same cancer type.

On the other hand, we implemented the recently published CBHIR framework based on Convolution Auto Encoder (CAE) into the skin data set to provide a comparison between the Skin-twins CBHIR approach and the CAE CBHIR algorithm. According to the comparison, the proposed Skin-twins CBHIR model overpassed the CAE CBHIR model with the difference of $67\%$ of precision, $68\%$ of Recall, and $67.5\%$ of F1Score higher amount at the top first retrieved images.

In this study, as far as the author is aware, for the first time, a CBHIR technique was proposed to tackle the challenges of grading Spitzoid Tumors of Uncertain Malignant Potential (STUMP) queries by retrieving top \textit{K} similar patches. Since STUMP tissues are highly challenging for pathologists, by retrieving top \textit{K} similar patches, pathologists can analyze the histopathological patterns based on their knowledge and grade the query STUMP more accurately and faster. As far as the importance of histopathological patterns in cancer diagnosis, we reported the Grad-CAM figures of the STUMP query and the retrieved patches to highlight the region of interest for the Skin-twins network for finding similarity. 

Based on the discussions in this paper and the high performance of the proposed CBHIR model in breast and skin cancer data sets with promising properties in histopathological diagnosis support across different cancer types, the CBHIR models can yield values in the daily workflow of hospitals in cancer diagnosis.

In further investigations, Federated Learning (FL) can enhance the performance of the CBHIR framework by training the model with richer data sets coming from different centers. This enhancement can assist pathologists in different medical centers to have more accurate cancer diagnoses with a generalized technique.
In addition, expanding the breast data set with different magnifications can provide a multi-magnification CBHIR model that can retrieve patches from different levels of magnifications. Consequently, pathologists can reach similar histopathological images at several levels of magnification.

Another future line can be dedicated to exploring histology foundation models within CBHIR systems. CBHIR models encompass both image and text data to enhance the retrieval of histology images based on their content.

\section{Author Contributions}
Zahra Tabatabaei: Conceptualization, Methodology, Analyzing, Writing- Original draft, Reviewing \& editing, Formal analysis, visualization.

Adrián Colomer: Supervision, Conceptualization, Methodology, and Review.

Javier Oliver Moll: Supervision, Conceptualization, Methodology.

Valery Naranjo: Supervision, Conceptualization, Methodology.

\section{Acknowledgment}
This study is funded by the European Union’s Horizon 2020 research and innovation program under the Marie Skłodowska-Curie grant agreement No. 860627 (CLARIFY Project \footnote{http://www.clarify-project.eu/}). The work of Adrián Colomer has been supported by Ayuda a Primeros Proyectos de Investigación (PAID-06-22), Vicerrectorado de Investigación de la Universitat Politècnica de València (UPV).

We would like to express our gratitude to Dr. Gonzalo Safont Armero and Daniel Fernández Gómez for their valuable assistance in this research.

\bibliographystyle{elsarticle-num}
\bibliography{cas-refs}

\end{document}